\documentclass[sigconf]{acmart}

\AtBeginDocument{%
 \providecommand\BibTeX{{%
    \normalfont B\kern-0.5em{\scshape i\kern-0.25em b}\kern-0.8em\TeX}}}

\usepackage{soul}
\usepackage{color}
\usepackage{wallpaper}
\usepackage{graphicx}
\usepackage{cleveref}
\usepackage{numprint}
\usepackage{paralist}
\usepackage{xspace}
\usepackage{amsmath}

\definecolor{orangeX}{rgb}{1,.5,0}
\definecolor{blueX}{rgb}{.2, .59, .88}
\definecolor{purpleX}{rgb}{.294118, 0, .509804}
\definecolor{greenX}{rgb}{.421, .578, .241}
\definecolor{bole}{rgb}{0.47, 0.27, 0.23}
\definecolor{mypink3}{cmyk}{0, 0.7808, 0.4429, 0.1412}
\definecolor{mygray}{gray}{0.6}
	
\newcommand{\ulmlong}{Ulm-Trier Social Stress\,}
\newcommand{\ulm}{\textsc{Ulm-TSST}}

\newcommand{\awe}{\textsc{RAAW}}
\newcommand{\awelong}{\emph{Rater Aligned Annotation Weighting\,}}

\newcommand{\ds}{\textsc{DeepSpectrum}}

\newcommand{\vgg}{\textsc{VGGish}}
\newcommand{\vggf}{\textsc{VGGface}}

\newcommand{\mtcnn}{\textsc{MTCNN\,}}
\newcommand{\bert}{\textsc{BERT\,}}

\newcommand{\eg}{e.\,g., }
\newcommand{\ie}{i.\,e., }

\newcommand{\cf}{{cf.\ }}

\begin{document}
\fancyhead{}
\title{
A Physiologically-Adapted Gold Standard \\ for Arousal during Stress}

\author{Alice Baird}
\affiliation{%
  \institution{Chair EIHW, University of Augsburg}
  \city{Augsburg, Germany}}

\author{Lukas Stappen}
\affiliation{%
  \institution{Chair EIHW, University of Augsburg}
  \city{Augsburg, Germany}}

\author{Lukas Christ}
\affiliation{%
  \institution{Chair EIHW, University of Augsburg}
  \city{Augsburg, Germany}}

  \author{Lea Schumann}
\affiliation{%
  \institution{Chair EIHW, University of Augsburg}
  \city{Augsburg, Germany}}
  
\author{Eva-Maria Meßner}
\affiliation{%
  \institution{KPP, University of Ulm}
  \city{Ulm, Germany}}

\author{Bj\"orn W.\ Schuller}
\affiliation{%
  \institution{GLAM, Imperial College London}
  \city{London, United Kingdom}}

\begin{abstract}
Emotion is an inherently subjective psychophysiological human-state and to produce an agreed-upon representation (gold standard) for continuous emotion requires a time-consuming and costly training procedure of multiple human annotators. 
There is strong evidence in the literature that physiological signals are sufficient objective markers for states of emotion, particularly arousal. In this contribution, we utilise a dataset which includes continuous emotion and physiological signals -- Heartbeats per Minute ($BPM$), Electrodermal Activity ($EDA$), and Respiration-rate  -- captured during a stress inducing scenario (Trier Social Stress Test). We utilise a Long Short-Term Memory, Recurrent Neural Network to explore the benefit of fusing these physiological signals with arousal as the target, learning from various audio, video, and textual based features. We utilise the state-of-the-art MuSe-Toolbox to consider both annotation delay and inter-rater agreement weighting when fusing the target signals. An improvement in Concordance Correlation Coefficient (CCC) is seen across features sets when fusing $EDA$ with arousal, compared to the arousal only gold standard results. Additionally, \bert-based textual features' results improved for arousal plus all physiological signals, obtaining up to .3344 CCC compared to .2118 CCC for arousal only. Multimodal fusion also improves overall CCC with audio plus video features obtaining up to .6157 CCC to recognize arousal plus $EDA$ and $BPM$. 
\end{abstract}



\maketitle

\section{Introduction}

Physiological and emotional responses can coincide during a stressful situation~\cite{duijndam2020physiological}, and the degree of correlation has shown to be dependent on factors including underlying psychological traits and states, \eg social desirability, or physiological dispositions, \eg brain morphology~\cite{campbell2012acute}.  For research on the discrepancy between physiological and self-reported emotional states see ~\cite{schwerdtfeger2004repressive}.  During a stress-inducing situation, heart-rate, and breath become varied~\cite{bernardi2000effects}, along with the voice~\cite{Baird2019TSST} (which is related strongly to perceived affect~\cite{eckland2019role}).  To this end, signals such as the Electrodermal Activity ($EDA$) -- described as a psycho-physiological indication of emotional arousal~\cite{caruelle2019use} -- correlate with an individual current perceived emotional state, specifically during high states of arousal, \eg during a competitive video game~\cite{Drachen2010}.

Within the field of affective computing, recognition approaches to predict continuous states of emotion frequently utilise the two-dimensional Circumplex Model of Affect~\cite{russell1980circumplex}, observing the arousal (activation) and valence (positivity) of perceived emotion. However, as emotion is a subjective state of being, multiple raters must continuously annotate, which is time-consuming and costly. Further to this, the method to obtain a robust agreed-upon signal from multiple raters (gold standard) remains an ongoing research question, with several methods available. For example, given the likelihood of disagreement, weighting annotators based on level of agreement can be applied using the Evaluator Weighted Estimator (EWE)~\cite{grimm2005evaluation}. Furthermore, annotator delay is not consistent per annotator, and so aligning rating with consideration to peaks is needed and Canonical Time Warping (CTW)~\cite{zhou2015generalized} can be applied in this case.

With this in mind, research into the fusion of physiological signals for use with perceived emotional signals is limited. Although physiological signals are utilised as features~\cite{dhall2020emotiw}, or extracted during particular tasks, to better target arousal~\cite{BairdSLT2021}, there has been minimal research on a combined physiological and perceived arousal gold standard. Recently, in the 2021 edition of the Multimodal Sentiment in-the-wild (MuSe) challenge, the signal of arousal was fused with $EDA$ and used as a prediction target for the \textit{MuSe-Physio} sub-challenge~\cite{stappen2021muse}. The baseline result from this was $0.3$ CCC stronger than the arousal only \textit{MuSe-Stress} sub-challenge when performing a late-fusion of audio and video-based features. Furthermore, the text-based features (typically less helpful for recognition of arousal) also improved through $EDA$ fusion with arousal, showing promise that has encouraged the authors to investigate further. However, to the best of the authors' knowledge, this was the first time that arousal and $EDA$ were fused in this manner, and there are no works that explore the fusion of arousal with other physiological signals such as respiration or heart rate (as $BPM$). 

To explore this idea further, in this contribution, we utilise the same dataset available through the MuSe Challenge, the \ulmlong dataset (\ulm), and explore the fusion of $EDA$, $BPM$, and respiration rate with arousal ratings. The Trier Social Stress Test (TSST), which the subjects of the \ulm{} dataset were undergoing, consists of a free-speech job interview scenario. Given this pseudo-professional setting, we specifically consider that utilisation of physiological signals (a more objective marker for arousal) will be of use here, as perceived arousal may be suppressed to make a better impression towards the interviewer~\cite{caruelle2019use}. For our experiments, we utilise the recently released MuSe-Toolbox\cite{stappen2021musetoolbox}\footnote{https://github.com/lstappen/MuSe-Toolbox}, to apply a novel approach \awelong (\awe) for signal fusion which considers both weighting and alignment of ratings to create a gold standard. We extract several multimodal features from audio, video, and textual transcriptions and apply a Long Short-Term-Memory-Recurrent Neural Network (LSTM-RNN) as a regressor, following a similar training procedure as outlined in the MuSe 2021 challenge.

\begin{table}[]
\centering
  \caption{Reported are the number (\#) of speakers and total duration of the data splits across Train, (Devel)opment and Test partitions for the sub-set of the \ulm{} dataset.}
 \resizebox{0.7\linewidth}{!}{
  \begin{tabular}{lrrrr}
    \toprule

     & Train & Devel. & Test & $\sum$ \\
    \midrule
    \#     & 33   & 9 & 11 & 53\\
    \midrule
    hh:mm:ss & 2:45:29 &0:45:32 & 0:55:33 & 4:26:36 \\
  \bottomrule
\end{tabular}
}
\label{tab:partitioning}
\end{table}

\section{The \ulm{} Dataset}

We make use of the \ulmlong dataset (\ulm) for our experiments, a multimodal dataset first utilised as a part of the MuSe 2021 challenge~\cite{stappen2021muse}. Within the \ulm{} dataset, the subjects are undergoing a TSST, which is a standardised and renowned experiment to induce states of stress, allowing for a controlled setting with high-quality data. The full \ulm{} dataset consists of recordings from 110 German-speaking individuals (ca.\ 10 hours), which are annotated for the continuous dimensions of emotion (valence and arousal). The continuous emotion ratings are recorded at a sampling rate of 2\,Hz and made by three annotators (obtaining an average inter-rater agreement for arousal of $.173$ Pearson's Correlation Coefficient). In addition, the modalities of audio, video, and text can be extracted from the dataset, as well as four captured physiological signals at a sampling rate of 1\,kHz: Electrodermal Activity ($EDA$), Electrocardiogram (ECG), respiration rate ($RESP$) as chest displacement during breath [-10:+10], and heart rate as beats per minute ($BPM$). For our experiments, we utilise a sub-set of the dataset as presented in the MuSe 2021 challenge, which was further processed, and reduced to 53 speakers.

The data is in a speaker-independent train, development, and test partitioning, with balanced speaker demographics across the partitions, \cf \Cref{tab:partitioning}. Before feature extraction, videos are cut from start to end of the TSST, and excluding participant names. We choose to use only $EDA$, $BPM$ and $RESP$ for the physiological signals, and each is down-sampled to 2\,Hz (to match the arousal ratings) and smoothed, applying a Savitzky–Golay filter, to reduce irrelevant, fine-grained artefacts in the signal. We exclude the ECG signal, as $BPM$ captures this activity at a higher level which is more optimal for the applied down-sampling.












\begin{table}[]
\centering
\caption{The mean ($\mu$) and standard deviation ($\pm$) for inter-rater agreement, as Pearson correlation coefficient (CC). Calculated during EWE after CTW alignment.}
\resizebox{0.6\linewidth}{!}{
\begin{tabular}{lll}
\toprule 
  CC              & $\mu$ & $\pm$    \\
\midrule
 $A_1$,$A_2$,$A_3$                          & .173 & .191 \\
 \midrule
 $A_1$, $A_2$, $EDA$                      & \textbf{.230} & .241 \\
 $A_2$,$A_2$ + $BPM$                      & .158 & .187 \\
 $A_2$,$A_2$ + $RESP$              & .108 & .134 \\
 \midrule
 $A_1$,$A_2$,$A_3$, $EDA$,$BPM$               & .119 & .155 \\
 $A_1$,$A_2$,$A_3$, $EDA$,$RESP$       & .088 & .120 \\
 $A_1$,$A_2$,$A_3$, $BPM$,$RESP$      & .070 & .097 \\
 \midrule
 $A_2$, $A_2$, $EDA$, $BPM$, $RESP$   & .127 & .123 \\
 \midrule 
 $EDA$, $BPM$, $RESP$ & .197 & .149 \\

\bottomrule
\end{tabular}
}
\label{tab:rater_ag}
\end{table}

\section{Experimental Settings}

To evaluate the benefit of fusing physiological signals with perceived arousal, we primarily conduct a series of continuous recognition tasks utilising various combinations of the three perceived arousal ratings with the $EDA$, $BPM$, and $RESP$ signals. 

\begin{figure}[h]
    \centering
    \includegraphics[width=0.8\linewidth]{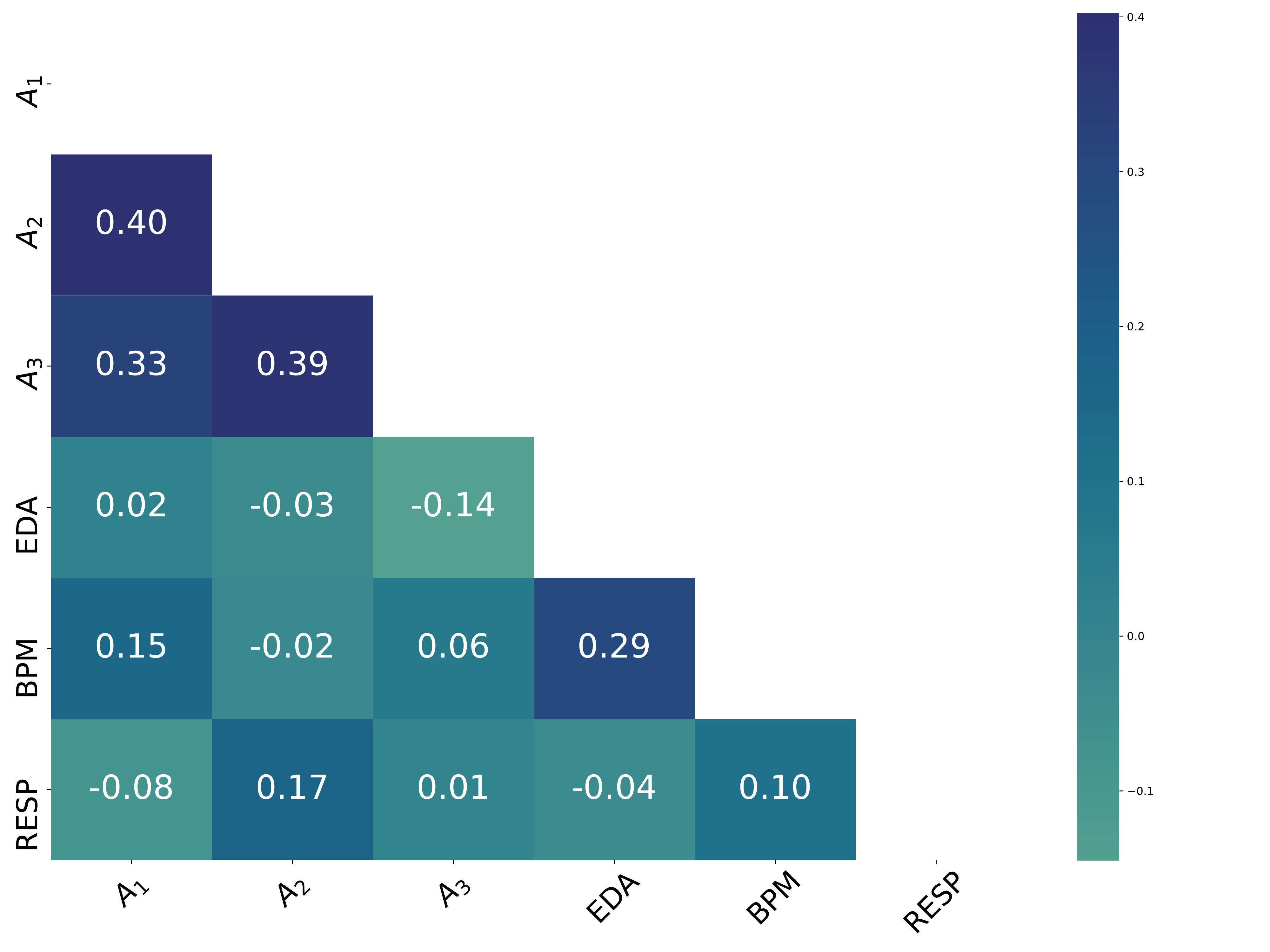}
    \caption{Correlation matrix between each individual signal across all speakers in the training set.}
    \label{fig:corr_matrix}
    \vspace{-0.5cm}
\end{figure}

\begin{enumerate}
\item $A_1$, $A_2$, $A_3$: Perceived arousal ratings only. From annotators one, two, and three.  
\item $A_1$, $A_2$, $EDA$: Arousal rater one ($A_1$) and Arousal rater two ($A_2$) plus EDA.  $A_1$ and $A_2$ are chosen as correlation is slightly higher for these signals compared to $A_3$, as shown in \Cref{fig:corr_matrix}. 
\item $A_1$, $A_2$, $BPM$. 
\item $A_1$, $A_2$, $RESP$. 
\item $A_1$,$A_2$,$A_3$, $EDA$,$BPM$.  
\item $A_1$,$A_2$,$A_3$, $EDA$,$RESP$.     
\item $A_1$,$A_2$,$A_3$, $BPM$,$RESP$.
\item $A_1$, $A_2$, $A_3$, $EDA$, $BPM$, $RESP$. 
\item $EDA$, $BPM$, $RESP$: Physiological signals only. 
\end{enumerate}
    \vspace{-0.4cm}

\subsection{Label Fusion Strategy}

We utilise a continuous annotator fusion technique \awelong (\awe), first presented in  \cite{stappen2021muse}. \awe{} is aimed at two challenges for gold standard creation of continuous emotion, a) annotator delay, and b) rater disagreement. Annotator delay is a typical issue with continuous ratings~\cite{ringeval2013introducing} and can be reduced through an explicit time-shift or through methods that automatically time-shift based on factors such as rater agreement, and in this instance, Canonical Time Warping (CTW) is applied~\cite{zhou2015generalized}. For weighting the annotators based on their agreement, the Evaluator Weighted Estimator (EWE)~\cite{grimm2005evaluation} is commonly applied for emotional gold standard~\cite{baird2019predicting}. 

For our experiments, we create six gold standards as shown in \Cref{tab:rater_ag} and above. For each, we are comparing to the perceived arousal only gold standard. There are two annotators ($A_1$, $A_2$) and a physiological signal. We explore the benefit of removing an annotator who is sub-optimal (in other words, the annotator with the lowest agreement). Where we apply two arousal raters with all physiological signals, we explore the advantage of using physiological signals to bring the rating in the gold standard up to five.

\subsection{Features}

For all features, we utilise the package provided from the MuSe 2021 challenge~\cite{stappen2021muse}. To reduce the scope of our experiments, we select the better performing features set from the baseline experiments. However, speech is strongly linked to perceived arousal, so we choose to use the two best performing feature sets. We offer a short description of critical points for the feature extraction process applied, for details \cf~\cite{stappen2021muse}.

\textbf{Audio:} We apply a six-second window size for the acoustic features. As a first step, the entire audio sequence is extracted from a given video. This file is then converted from stereo to mono with a sampling of 16\,kHz, 16\,bit, and then normalised to -3\,dB. For \ds{}, we keep the default settings for extraction to obtain a 4\,096-dimensional feature set. For \vgg{}, by aligning the frame and hop size to the annotation sample rate, we extract a 128-dimensional \vgg{} embedding vector every 0.5\,s from the underlying log spectrograms.

\textbf{Video: }Given the human nature of this task, video-based features focus on the face, although it may be beneficial to explore gesture-based features more specifically in further research. The \mtcnn~\cite{zhang2016mtcnn} is used to extract faces. \mtcnn is pre-trained on the datasets WIDER FACE~\cite{DBLP:journals/corr/YangLLT15b} and CelebA~\cite{liu2015faceattributes}. The \mtcnn extractions are used as inputs for \vggf{}. \vggf{}~\cite{Omkar2015recognition} is aimed at the extraction of general facial features and is based on detaching the top-layer of a pre-trained version of the deep CNN referred to as VGG16~\cite{simonyan2014very}. This results in a 512 feature vector output. 


\textbf{Test: }For extracting the text features from the transcripts, a pre-trained Transformer language model \bert~\cite{devlin2019bert}, is used. We obtain word-level features from the sum of the last four \bert encoder layers resulting in a 768-dimensional feature vector for each word analogous to~\cite{sun2020multi}. This data contains exclusively German speech. For this reason, the \bert pre-trained on German texts\footnote{\url{https://deepset.ai/german-bert}} is applied. 


\textbf{Alignment: }The label-aligned features are made available from the MuSe challenge. These include the same frame rate as the provided label arousal labels. For the textual features of the \ulm{}, as they are based on manual transcripts of the videos, the Montreal Forced Aligner (MFA)~\cite{mcauliffe2017montreal} tool is applied to obtain word-level timestamps.

\begin{table*}[]
\small
\caption{Reporting Concordance Correlation Coefficient (CCC) results for prediction of nine combinations of perceived arousal and physiological-arousal signals on the devel(opment) and test partitions. Utilising (V)ision: \vggf, (A)udio: \ds, \vgg, and (T)ext: \bert. Reporting the best result from hyperparameter optimisation, as well as  reporting the mean ($\mu$) across all feature sets for a given signal. Best test scores are emphasised.}
\resizebox{\linewidth}{!}{
\begin{tabular}{l r r r r r r r r r r r r r r r r r r}
\toprule
                 \textit{Percieved} & \multicolumn{2}{c}{\boldmath{$A_1$,$A_2$,$A_3$}} & \multicolumn{2}{c}{\boldmath{$A_1$,$A_2$}} & \multicolumn{2}{c}{\boldmath{$A_1$,$A_2$}} & \multicolumn{2}{c}{\boldmath{$A_1$,$A_2$}} & \multicolumn{2}{c}{\boldmath{$A_1$,$A_2$,$A_3$}} & \multicolumn{2}{c}{\boldmath{$A_1$,$A_2$,$A_3$}} & \multicolumn{2}{c}{\boldmath{$A_1$,$A_2$,$A_3$}} & \multicolumn{2}{c}{\boldmath{$A_1$,$A_2$}} &   \\
                  \textit{Physiological} & \multicolumn{2}{c}{} & \multicolumn{2}{c}{\textbf{$EDA$}} & 
                 \multicolumn{2}{c}{\textbf{$BPM$}} & 
                 \multicolumn{2}{c}{\textbf{$RESP$}} & \multicolumn{2}{c}{\textbf{$EDA$,$BPM$}} & \multicolumn{2}{c}{\textbf{$EDA$,$RESP$}} & \multicolumn{2}{c}{\textbf{$BPM$,$RESP$}} & \multicolumn{2}{c}{\textbf{$EDA$,$BPM$,$RESP$}} & \multicolumn{2}{c}{\textbf{$EDA$,$BPM$,$RESP$}}   \\

     CCC            & Devel          & Test           & Devel           & Test           & Devel           & Test           & Devel            & Test           & Devel             & Test             & Devel             & Test              & Devel             & Test              & Devel               & Test      & Devel & Test          \\
                 \midrule
\vggf          & .3025         & .3813         & .3216          & .3959         & .4805          & .3771         & .1869           & \textbf{.3745         }& .3694            & .4062           & .3995            & .3941            & .3637            & .4306            & .4704              & .4707           & .5679    & \textbf{.5838} \\
\ds     & .2826         & .3060          & .3366          & .4031         & .1649          & .2327         & .0382           & .0977         & .3089            & .3861           & .1841            & .3807            & .2527            & .2046            & .3683              & .3832         & .4189                    & .5157            \\
\vgg           & .2127         & .2856         & .3493          & .4210          & .3156          & .3313         & -.0079          & .1716         & .4851            & .5164           & .0901            & .3985            & .2689            & .3649            & .5161              & .4712          & .3197                    & .4613          \\
\bert             & .1341         & .2118         & .2431          & .2402        & .0567          & .1037         & .1063           & .1802         & .1999            & .0542           & .2733            & .2393           & .1210             & .0922            & .3568              & .3344        & .2909                    & .3842          \\
\midrule
\multicolumn{17}{c}{\textbf{Late-Fusion}} \\
\midrule
A + V      & .4638         & \textbf{.5062         }& .4506          & \textbf{.5103 }        & .4640           & .3889         & .3196           & .3108         & .5666            & \textbf{.6157 }          & .3630             & \textbf{.3947}            & .4722            & \textbf{.4432 }           & .6674              & .5025           & .5030                    & .5728             \\
A + T       & .3240          & .3841         & .3821          & .3470          & .3044          & .3205         & .1396           & .2032         & .5089            & .3677           & .3249            & .1777            & .3295            & .2817            & .5570               & .4357             & .4175                    & .5586   \\
V + T       & .2526         & .4668         & .3442          & .4213         & .4735          & \textbf{.4202 }        & .3443           & .2871         & .4839            & .3783           & .3836            & .2301            & .3738            & .3881            & .5916              & \textbf{.5355 }         & .4386                    & .5594      \\
A + V + T & .3476         & .4965         & .4186          & .4987         & .4458          & .4104         & .3811           & .3036         & .5895            & .4596           & .4028            & .3470             & .4086            & .4230             & .6669              & .5055      & .4623                    &.5639   \\
\midrule
\midrule
$\mu$ of All & --& 	.3798		&--&.4047	&--&	.3231	&--&	.2411	&--&	.3980	&--&	.3203	&--&	.3285&--&		.4548  & -- & \textbf{.5250}\\

\bottomrule
\end{tabular}}
\label{tab:results}
\end{table*}


\begin{table}[]
\small
\caption{Mean ($\mu$) and standard deviation ($\pm$) for a selection of test results given in \Cref{tab:results}, which include either the EDA, BPM, or RESP signal.}
\resizebox{\linewidth}{!}{
\begin{tabular}{l r r r r r r r}
\toprule
                 &  \boldmath{$A_1$,$A_2$,$A_3$} & \multicolumn{2}{c}{\textbf{inc. $EDA$ }} & \multicolumn{2}{c}{\textbf{inc. $BPM$ }} & \multicolumn{2}{c}{\textbf{inc. $RESP$}} \\
                 &              & $\mu$            & $\pm$             & $\mu$            & $\pm$             & $\mu$             & $\pm$             \\
                 \midrule
\vggf          & .3813       & .4167          & .0364          & .4212          & .0396          & .4175           & .0424          \\
\ds     & .3060        & .3883          & .0101          & .3017          & .0965          & .2666           & .1402          \\
\vgg           & .2856       & .4518          & .0527          & .4210          & .0872          & .3516           & .1279          \\
\bert             & .2118       & .2170          & .1174          & .1461          & .1273          & .2115           & .1018          \\
\midrule
\multicolumn{8}{c}{\textbf{Late-Fusion}} \\
\midrule
A + V      & .5062       & .5058          & .0903          & .4876          & .0972          & .4128           & .0810          \\
A + T       & .3841       & .3320          & .1096          & .3514          & .0663          & .2746           & .1162          \\
V + T       & .4668       & .3913          & .1263          & .4305          & .0722          & .3602           & .1339          \\
A + V + T & .4965       & .4527          & .0733          & .4496          & .0427          & .3948           & .0888   \\
\bottomrule
\end{tabular}
}
\label{tab:mean}
\end{table}

\begin{figure}
    \centering
    \includegraphics[trim={0 0 0 1.1cm},clip,width=0.9\columnwidth]{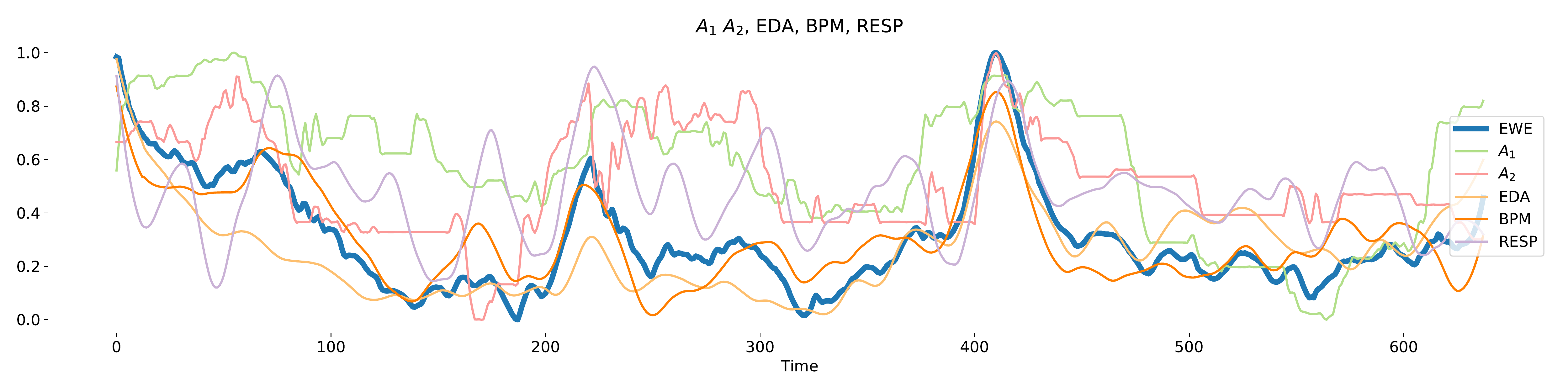}\\
    \includegraphics[trim={0 0 0 1.1cm},clip,width=0.9\columnwidth]{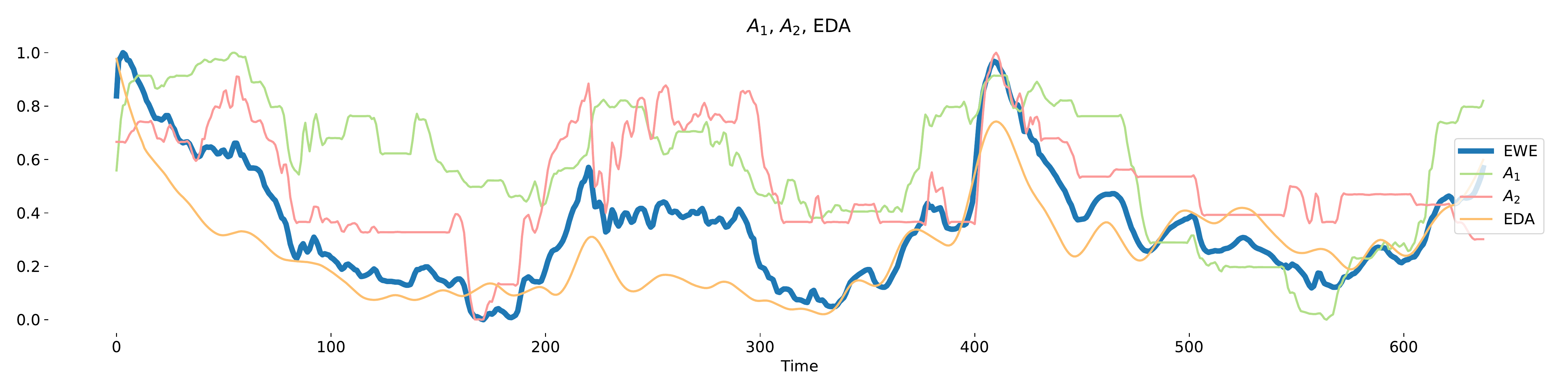} \\
    \includegraphics[trim={0 0 0 1.1cm},clip,width=0.9\columnwidth]{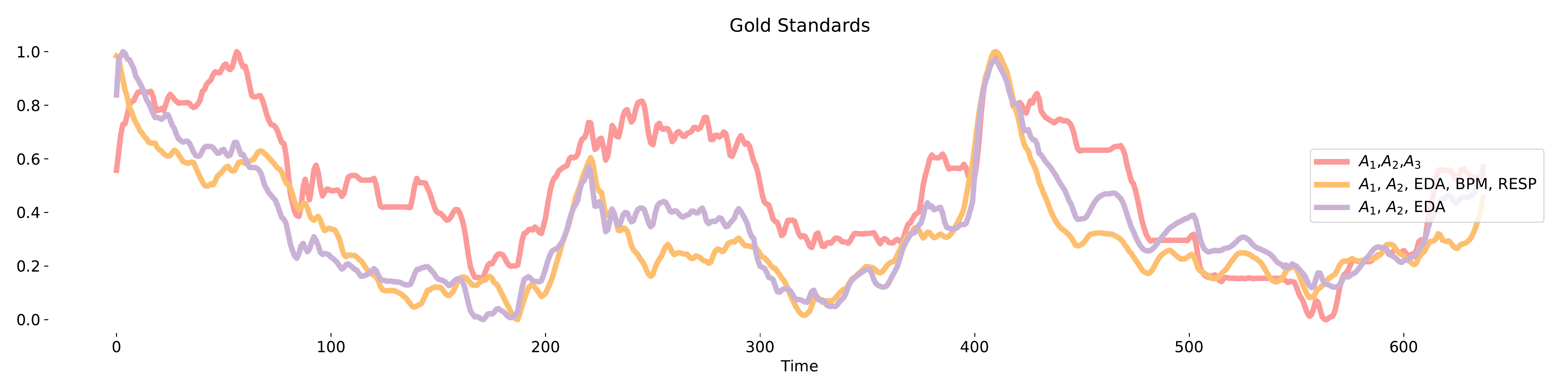} \\
    \caption{An example for subject \#\,9 for three of the gold standards used in these experiments. (Upper) $A_1$, $A_2$ + $EDA$, $BPM$ and $RESP$ ($\pm$: 0.203). (Middle) $A_1$, $A_2$ + $EDA$ ($\pm$: 0.217). (Lower) comparison of the above gold standards including arousal only ($\pm$: 0.241).}
    \label{fig:signalcompare}
\end{figure}

\subsection{Regressor: LSTM-RNN}

Given the time-dependent nature of this task, we utilise the same LSTM-RNN based architecture as applied for the baseline of the MuSe 2021 Challenge\footnote{\url{https://github.com/lstappen/MuSe2021}}. Extensive hyperparameter optimisation is applied, and the extracted feature sequences are input into a uni- and bi-directional LSTM-RNN with a hidden state dimensionality of $h = \{32, 64, 128\}$, to encode the feature vector sequences. We test different numbers of LSTM-RNN layers $n = \{1, 2, 4\}$, and search for a suitable learning rate $lr=\{0.0001, 0.001, 0.005\}$. For further detail of the architecture applied \cf~\cite{stappen2021muse}. In the training processes, the features and labels of every input video are further segmented via a windowing approach~\cite{sun2020multi}. As in the MuSe Challenge, a window size of 300 steps (150 seconds) and a hop size of 50 steps (25 seconds) is used.

\subsection{Feature Fusion}
To observe the benefits of multimodal approaches, we apply a  decision-level (late) fusion to evaluate the co-dependencies of the modalities. The experiments are restricted to the best performing features from each modality only. For decision-level fusion, separate models are trained individually for each modality. 
The predictions of these are fused by training an additional LSTM-RNN model as described above. For all tasks, we apply a unidirectional version with $lr = 0.001$, $h = 64$, and $n = 4$.

\section{Discussion of Results}
To explore the benefit of physiological-based arousal and perceived arousal fusion, the extensive results for the computational prediction experiments conducted are given in \Cref{tab:results}, and \Cref{tab:mean}. As an evaluation metric for these experiments, CCC is employed, as is typical for emotion recognition tasks, and to better compare to the initial baseline results obtained using the \ulm{} dataset~\cite{stappen2021muse}. 

For the results in \Cref{tab:results}, we see that the perceived arousal only ($A_1$-$A_3$) score is strong, particularly from a multimodal approach where at best $0.5062$ CCC is achieved on the test set, from late-fusion of audio and video-based features. However, looking at the uni-modal approaches for $A_1$-$A_3$, as we expected given the pseudo-professional scenario of the TSST, audio-only features capture the perceived arousal to lesser degree compared to \vggf{}. Furthermore, as is typical for arousal prediction tasks, the uni-modal textual features perform worst, obtaining $0.2118$ CCC on the test set. 

As we move along the table (\cf \Cref{tab:results}) to the right, we see in general a slight improvement across features when incorporating a physiological signal. Of interest, we see a ca.\ $.3$ CCC improvement for \bert features when utilising the $EDA$ signal. Typically, perceived arousal is a challenging task for textual-based features, as seen from the perceived arousal baseline. However, at best for \bert features when predicting the combined $A_1$, $A_2$, $EDA$, $BPM$, $RESP$ signal, we obtain $.3344$ CCC, which is $.1$ above the $A_1$-$A_3$ baseline. When observing the mean across experiments, including $EDA$~\Cref{tab:mean}, we confirm that $EDA$ is the strongest physiological signal for the \bert features. 

We also see the audio features obtain a more robust result than \vggf{} when utilising $EDA$, suggesting that the behaviour of $EDA$ is present in the voice, making this gold standard more attainable for the speech-based features. Similar behaviour for $BPM$ and $RESP$ fusion results are obtained; however, this is not as consistent as $EDA$ results, as we can see through the more consistent mean results in~\Cref{tab:mean}. Furthermore, there are lower results for the $A_1$, $A_2$ $BPM$, and $A_1$, $A_2$ $RESP$ results, compared to the $A_1$-$A_3$ baseline. 

For audio features, all gold standard approaches, which include $EDA$ report an improvement, and up to $.4712$ CCC is obtained by \vgg{} features, where all physiological signal are utilised. In \Cref{fig:signalcompare}, we can see that the two physiological-adapted gold standards follow a similar trend to the arousal baseline gold standard. However, there is a slightly reduced standard deviation for this example, with $0.24$ for $A_1$-$A_3$ compared to $0.22$ for $A_1$, $A_2$, $EDA$, and 0.203 for the $A_1$, $A_2$, $EDA$, $BPM$, $RESP$ signal. This may suggest that better results are obtained from a smoothing effect when the perceived arousal is fused with physiological signals. 

\begin{figure}
    \centering

    \includegraphics[
    width=0.48\columnwidth]{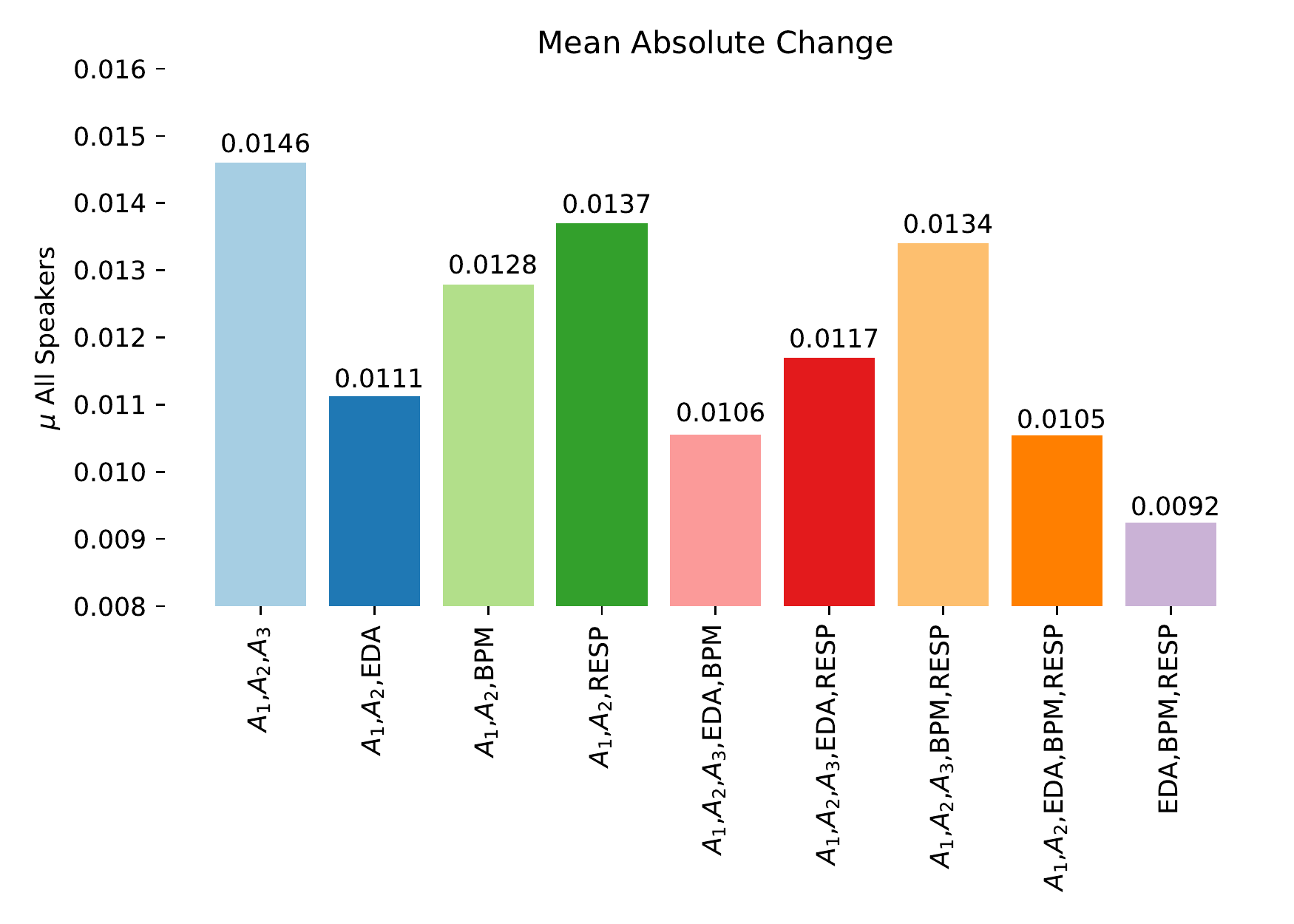}
\vspace{0.1cm}
    \includegraphics[
    width=0.48\columnwidth]{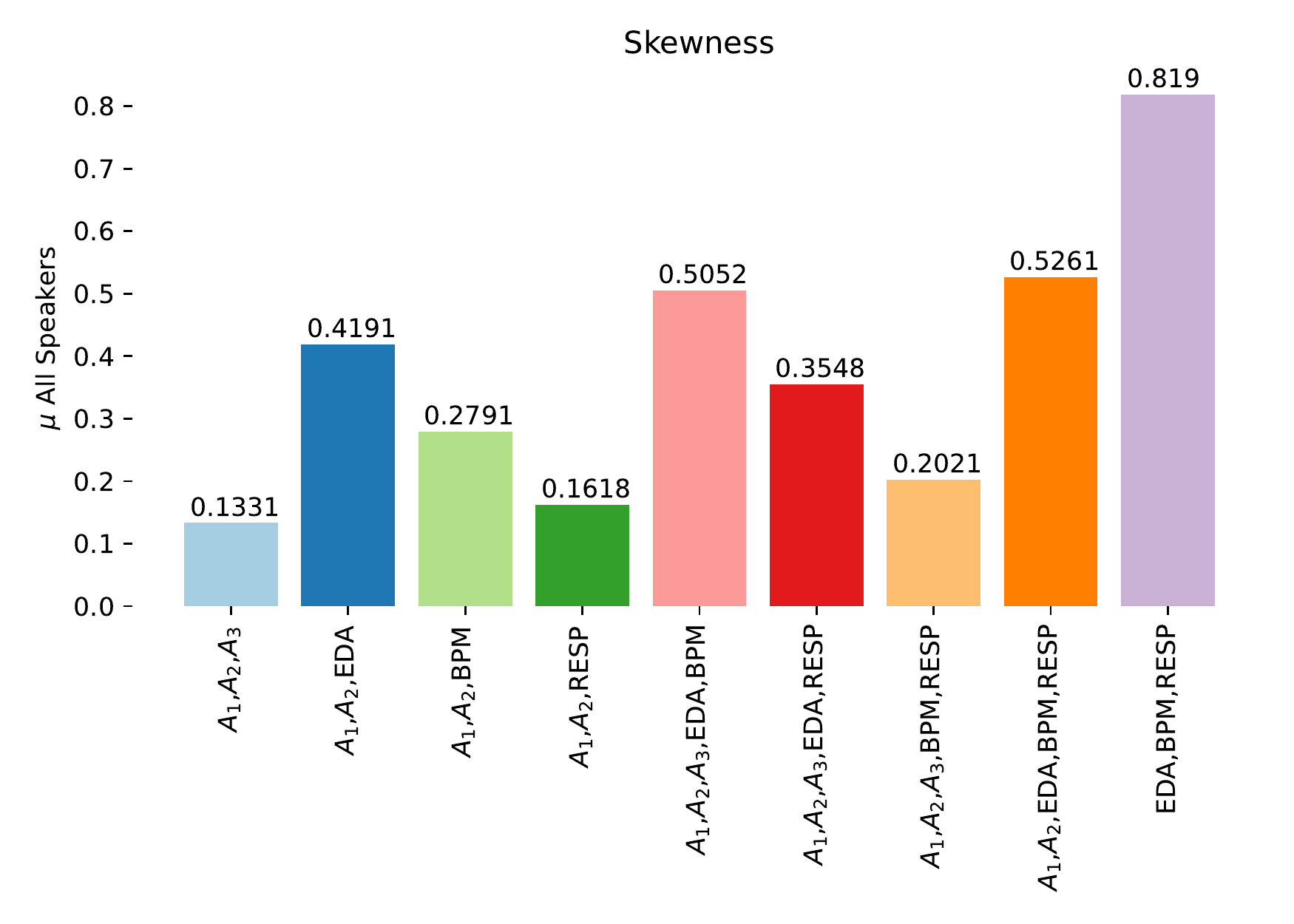}

    \caption{The mean absolute change and Skewness for the mean ($\mu$) of all speaker from each gold standard signal.}
    \label{fig:skew}
    \vspace{-20pt}
\end{figure}

To further analyse this smoothing effect, we see in \Cref{tab:results} that the results are consistently higher than the arousal only baseline when utilising the physiological only ($EDA$-$RESP$) signal. With a standard deviation of around $0.157$ for the same example in \Cref{fig:signalcompare}, we do lean more toward this being a factor in results improvement. We additionally extract the mean absolute change (MAC), and skewness from each of the gold standard (\Cref{fig:skew}) across all speakers. Although further investigation should be done here, we see that there is an inherent difference in the MAC from $A_1$-$A_3$, and $EDA$-$RESP$, which is mirrored by the skew of the signals' distribution. Of promise, and perhaps opposing the smoothing effect, none of the physiological signal results obtains higher than the best result when fusing with perceived arousal, \ie $.6157$ CCC from $A_1$, $A_2$, $EDA$, $BPM$ with audio and video feature fusion. This leads us to consider that further investigation on this topic may be fruitful -- particularly, as we do not see any reduction in results from physiological-adapted arousal fusion.

\vspace{-0.1cm}

\section{Conclusion}

To explore recognition of internal emotional states and improve the current state-quo for emotion-based gold standard creation, for the first time in this work, we explored the prediction of fused physical-based arousal with perceived arousal. We utilised the \ulm{} dataset, which offers a scenario in which stress is induced, and ultimately a testing bed for states of arousal. Findings have shown that in most cases, the $EDA$ signal can improve recognition of arousal, specifically textual based features, and aid acoustic features, which the less aroused speech behaviours may have challenged. There was less of an improvement from $BPM$ or $RESP$ signals alone, however, when fused with $EDA$, various feature sets did see improvements, with the best score obtained from a fusion of perceived arousal with $BPM$ and $EDA$ of up to $.6157$ CCC with late-fusion of audio and video features. One observation consistent throughout the experiments was the reduction in the standard deviation for the gold standard with physiological signals. Results seem to indicate that this aided the learning process, and it would be of interest to explore this more deeply in future work. 

As the original sampling rate available for the physiological signals was 1\,kHz, it would also be interesting to explore this in more detail. For example, the Savitzky–Golay filtering and extreme downsampling may have caused vital information loss. Furthermore, validation through other datasets would be optimal for exploring more deeply the combinations of signals and perhaps exploring a segmentation approach that is more suitable to the physiological signal. Another next step is to explore the valence dimension in this context. However, there is less literature supporting the manifestation of valence via physiological signals, so we have excluded it from our current research. 

\vspace{-0.1cm}

\section{Acknowledgments}
This project has received funding from the DFG's Reinhart Koselleck project No.\ 442218748 (AUDI0NOMOUS).

\bibliographystyle{unsrt}  
\bibliography{references}  

\end{document}